\documentclass{article}

\usepackage{arxiv}

\usepackage[utf8]{inputenc} 
\usepackage[T1]{fontenc}    
\usepackage{hyperref}       
\usepackage{url}            
\usepackage{booktabs}       
\usepackage{amsfonts}       
\usepackage{nicefrac}       
\usepackage{microtype}      
\usepackage{lipsum}

\usepackage[figuresright]{rotating}
\usepackage[caption=false, font=footnotesize]{subfig}

\usepackage{amssymb}
\usepackage{amsthm}
\usepackage{amsmath}
 \usepackage[scientific-notation=true]{siunitx} 
\usepackage[normalem]{ulem}   

\usepackage[ruled,vlined,linesnumbered]{algorithm2e}
\usepackage{setspace}

\usepackage{booktabs}
\usepackage{multirow}
\usepackage{complexity}

\usepackage{array, longtable, tabularx}
\usepackage{pdflscape}

\usepackage{graphicx}
\usepackage[super]{nth}

\usepackage{xcolor}

\usepackage{lineno,hyperref}
\modulolinenumbers[5]

\title{Multiple regression techniques for modeling dates of first performances of Shakespeare-era plays}

\author{
  Pablo Moscato\\
  College of Engineering, Science and Environment,\\ 
  The University of Newcastle,\\
  University Drive, Callaghan, NSW 2308, Australia\\
  \texttt{Pablo.Moscato@newcastle.edu.au} \\
  \And Hugh Craig\\\
  School of Humanities and Social Science,\\ 
  The University of Newcastle,\\
  University Drive, Callaghan, NSW 2308, Australia\\
  \texttt{{Hugh.Craig@newcastle.edu.au}}
   \And
 Gabriel Egan \\
  School of Humanities, \\
  De Montfort University,\\
  The Gateway, Leicester, UK, LE1 9BH\\
  \texttt{gegan@dmu.ac.uk} \\
   \And
   Mohammad Nazmul Haque \\
    College of Engineering, Science and Environment,\\ The University of Newcastle,\\
    University Drive, Callaghan, NSW 2308, Australia \\
   \texttt{Mohammad.Haque@newcastle.edu.au} \\
   \And
   Kevin Hunag\\
   California Institute of Technology\\
   1200 E California Blvd, Pasadena, CA 91125, US\\
   \texttt{khuang@caltech.edu} \\
   \And
   Julia Sloan\\
   California Institute of Technology, \\
   1200 E California Blvd, Pasadena, CA 91125, US\\
   \texttt{jsloan@caltech.edu} \\
   \And
   Jon Corrales de Oliveira\\
   California Institute of Technology, \\
   1200 E California Blvd, Pasadena, CA 91125, US\\
   \texttt{jonco@caltech.edu} \\
   
}

\begin{document}
\maketitle

\begin{abstract}
The date of the first performance of a play of Shakespeare's time must usually be guessed with reference to multiple indirect external sources, or to some aspect of the content or style of the play. Identifying these dates is important to literary history and to accounts of developing authorial styles, such as Shakespeare's. In this study, we took a set of Shakespeare-era plays (181 plays from the period 1585--1610), added the best-guess dates for them from a standard reference work as metadata, and calculated a set of probabilities of individual words in these samples. We applied 11 regression methods to predict the dates of the plays at an 80/20 training/test split. We withdrew one play at a time, used the best-guess date metadata with the probabilities and weightings to infer its date, and thus built a model of date-probabilities interaction. We introduced a memetic algorithm-based Continued Fraction Regression (CFR) which delivered models using a small number of variables, leading to an interpretable model and reduced dimensionality. An in-depth analysis of the most commonly occurring 20 words in the CFR models in 100 independent runs helps  explain the trends in linguistic and stylistic terms. The analysis with the subset of words revealed an interesting correlation of signature words with the Shakespeare-era play's genre.
\end{abstract}


\section{Introduction and motivation for the study}

In 1778, a century and a half after Shakespeare's death in 1616, the scholar Edmond Malone published the first attempt to give dates to Shakespeare's plays and to place them in chronological order~\cite{malone1778attempt}. Malone relied on allusions to the plays in documents surviving from Shakespeare's time and on evidence from the early printed editions. He admitted to many doubts and uncertainties about his suggested dates and ordering and the debate has continued unabated since.

Shakespeare's plays and those of his contemporaries were performed and printed in an era when little attention was paid to recording dates for posterity. The focus of the theatre was commercial and theatrical, rather than literary or archival. Since stage performance was paramount, and audiences in the theatre paid almost all the bills, with printed versions and income from commissions to perform at court accounting for only a fraction of revenue, drama was in large part an ephemeral art form. Many plays, perhaps the majority, have been lost and the documentation for those that survive is incomplete.

Over the years various new kinds of evidence about dating have been added to supplement what can be gleaned from the documentary record. In the latter part of the nineteenth century Frederick G. Fleay argued that changes in Shakespeare's versification were a useful guide to chronology~\cite[pp.\ 122--38]{fleay1876shakespeare} and this was taken up and extended by twentieth-century researchers~\cite{Gray:1931,langworthy1931verse,wentersdorf1951shakespearean,oras1960pause} and continues in the twenty-first~\cite{Bruster-Smith:2016}. The editors of the 1987 Oxford Shakespeare used changes in the incidence of colloquialisms in the dialogue of the plays as an index of the order of the plays~\cite[pp.\ 69--144]{wells1987william}. MacDonald P. Jackson drew on the progressive decline in the length of speeches in Shakespeare as a marker of chronology~\cite[Table\ 25.4]{Jackson:2007a,kar60210}. 

The language of the plays more broadly has also been analyzed for clues to dating. Eliot Slater introduced shared rare words as evidence of links between Shakespeare plays written at about the same time~\cite{slater1975shakespeare,slater1988problem}. Jackson has taken this up in various Shakespeare chronology studies, using larger text collections to calibrate rarity~\cite{jackson1978linguistic,jackson_2006,jackson2011deciphering,jackson2014vocabulary,jackson2018vocabulary}. 
\cite{waller1966use} looked at incoming and outgoing word forms like 'does' and 'doth' and~\cite{brainerd1980chronology} collected  a set of words which appeared to vary in incidence with date in Shakespeare plays. \cite{craig_2013} sought markers of change over time in very common and rarer words by comparing the language of sets of plays by Shakespeare's contemporaries as well as Shakespeare from different eras.

In the present paper we return to the question of chronology, sample widely in Shakespeare-era plays, focus on language features, and aim to construct a state-of-the-art model of change over time based on word probabilities. We take advantage of advances in multivariate regression to build an accurate model with a small set of variables, thus limiting dimensionality and simplifying the task of understanding the mechanism in linguistic and stylistic terms. We estimate the reliability of our model both for training and for test data.

For students of these plays, the date of first performance is generally the most informative for a chronology. The impact of the work on audiences and other writers begins with the first performance. In a fast-moving, competitive commercial environment, it can be assumed that composition occurred close to the date of first performance. The date of composition may seem a more logical starting point, but it would have to take account of spans of time:  the  first  creative  impulse  might  be many years prior to its realization, the work might be drafted and then put  aside for years, and so on. The date of printing, though usually easy to ascertain, is not as  useful as the date of first performance since it is often clearly widely separated from the date of the first performance, as with the eighteen plays associated with Shakespeare which were first printed in the Folio of $1623$, seven years after Shakespeare’s death in $1616$.

The date of first performance can sometimes be fixed with certainty because a performance is mentioned, and mentioned as the first, in an official document, a reliable personal diary, or in a printed work. There are also some records kept by theatre managers which helpfully record the date of performances. In most cases, however, the best we can do is determine a date we can be reasonably certain is the earliest possible, another which we can be reasonably certain is the latest possible, and then a single year which can be hazarded as a best guess. 

If we could devise a method to assign a date of first performance from internal evidence, using the evolution of style, for instance, as a continuum along which to place a given work, this would provide firmer foundations for the literary history of the drama of the time. The study in this paper is the first to offer a prediction of Shakespeare-era play dates based on internal evidence, validated by test samples, and extending beyond Shakespeare works.

\section{Materials and Methods}

\subsection{Dataset of 285 plays}

\begin{figure}
\centering 
\includegraphics[width=0.75\textwidth]{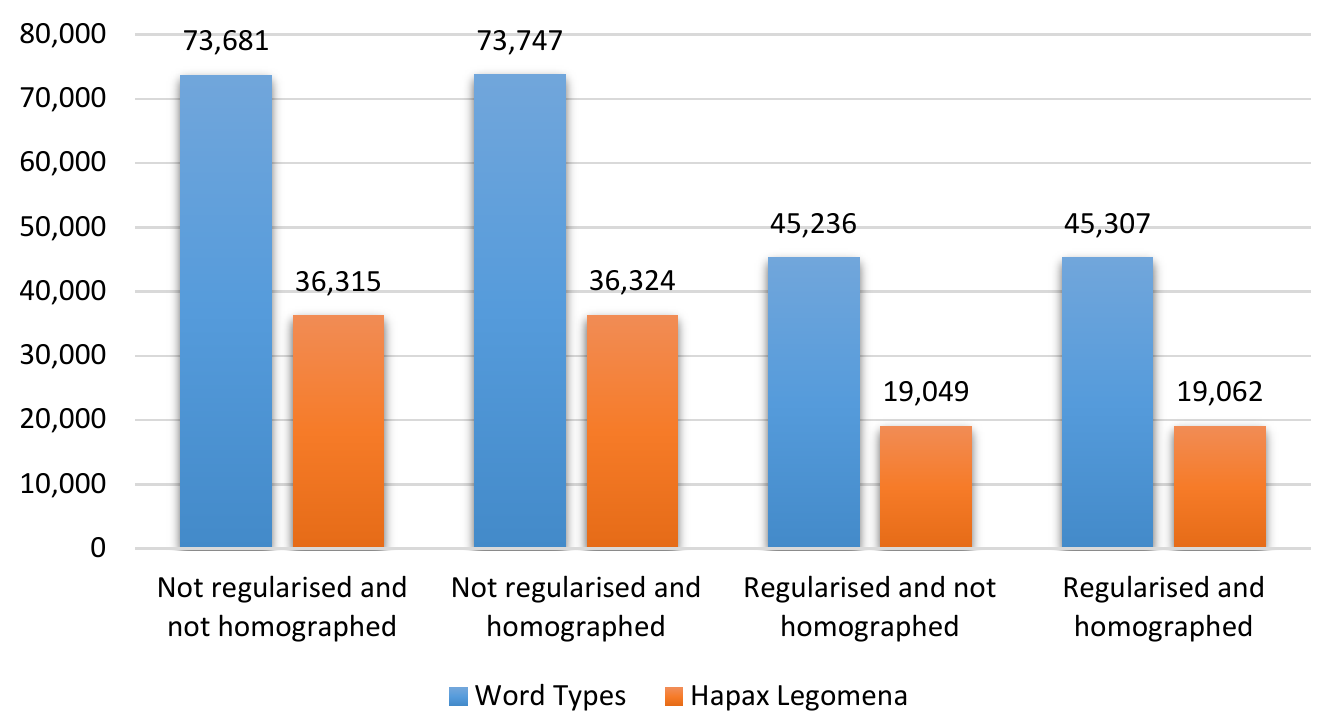}
\caption{Word Types and hapax hegomena in 143 Plays. In these plays, a subset of the corpus, the mark-up of the text allows us to retrieve the state of the text before regularization and the tagging of homographs. Marking a select list of homographs makes only a small difference in totals. Hapax legomena (word types represented in the corpus by only one instance) are half the total in unregularized  text, and less than half in regularized text. After regularization, the remaining word types are three-fifths of the original total and the remaining hapax legomena are around half the original total.}\label{fig:Word-types-and-hapax-legomena}
\end{figure}

For this study, we have used a collection of 285 English plays from the sixteenth and seventeenth centuries assembled by Hugh Craig and Gabriel Egan for another as yet unpublished project on stylistic aspects of Folio versions of Shakespeare plays. This collection is a selection from the surviving printed and manuscript play versions from the period, with a bias towards original plays which had been performed by a professional company, rather than translations, plays written for  school, university and Inns of Court productions, or for readers as opposed to live audiences. There are 223 plays attributed to a sole author; 53 different playwrights are represented in this group. In addition, there are 26 multi-author plays and 36 plays of uncertain authorship. Craig and Egan took the earliest printed version as the basis for the machine-readable texts, except where this version is a manifestly corrupt one, as when it is obviously missing large sections or has manifestly garbled content. In some cases, we included alternative versions of the plays bearing in mind the scholarly interest in alternative Shakespeare versions in particular. Using early versions is preferable to using more recent ones since they have not been subject to modern editing, but this choice means that the spelling is variable. Spelling was not standardized in England until the late seventeenth century and before that multiple variant spellings were tolerated -- perhaps hardly noticed -- even within a single work. This creates difficulties for statistical methods based on word counting. The proliferation of variant spellings in these works is considerable, and confounds the expectations of anyone used to modern standardized spelling. {\cite{grazia1993materiality} found fourteen different spellings of the word `one' in printed works from this period}. The latitude in manuscript works is wider still. Jackson found sixteen spellings in a short manuscript which were not repeated anywhere in a large corpus of sixteenth and seventeenth century printed works~\cite{jackson2007hand}. Many words that are distinct in modern English overlapped in spelling in early modern English. The spelling {`weeke'}, for instance, was used for the different senses {`weak'}, {`week'} and {`wick'}; the forms {`travel'} and {`travail'}, {`hart'} and {`heart'}, and {`metal'} and {`mettle'} were interchangeable~\cite[pp.\ 37--38]{Craig_2010}. For these reasons  we modernized and standardized the spelling in the texts, using the Variant Detector (VARD) 2 software\footnote{\url{http://ucrel.lancs.ac.uk/vard}}~\cite{baron2008vard2,Baron_Rayson_Archer_2009} which offers assistance by prompting the user with probable modern equivalents and allowing global changes where the user feels confident there is only one possible modern equivalent for all the instances of a variant spelling in a work. (Figure~\ref{fig:Word-types-and-hapax-legomena}, below, shows the compression in word types in a section of the corpus that this step in pre-processing caused.)

We also marked up the works in the Text Encoding Initiative (TEI) P4 format, which uses a customized set of XML tags chosen to suit textual matter, so that stage directions, speech prefixes, prefaces, dedications and other non-dialogue material is identified and can be programmatically excluded from word counts. 

The standardization of spelling and parsing of text into dialogue and other materials is laborious, and no comprehensive collection of texts prepared in this way is available, so $285$ texts is a large collection compared to other studies apart from those using raw texts based on the Optical Character Recognition of digitized page images and machine-only standardization and parsing, where a considerable volume of error is encountered~\cite{HillHengchen:10.1093/llc/fqz024}.

The largest comparable open access manually curated collection of Shakespeare-era plays known to us is the first two components of the Enhanced Shakespeare Corpus (ESC). These include 36 Shakespeare plays and 46 plays by other authors. The ESC has a third, much more extensive component, including many more plays, as well as works in other text types, but the spelling standardization in this part was carried out programmatically, and those responsible warn that this produces a lower level of reliability than manual standardization by humans~\cite{Sean:2019}.

Using XML tags, we also marked a subset of words for part of speech so as to separate different uses of some grammatical words. These tags enable us to count instances of ``that'' as either conjunctions (as in ``she said that she would''), relatives (``the book that I left''), or demonstratives (``see that sword''), for instance. In all 19 grammatical words are marked in this way, yielding 48 separate word-forms for counting. The effect of this separation of some homographs, along with the impact of standardizing spelling, is illustrated in Figure~\ref{fig:Word-types-and-hapax-legomena}.

\subsubsection{Metadata}

We use a standard reference work, the multi-volume \textit{Catalogue of British Drama 1533--1642}~\cite{Wiggins:2012}, for dates of the first performance. This work offers a single best guess date for first performance for each play based on the latest theatre-historical investigations.

\subsubsection{Data availability}

After acceptance of this manuscript for publication, the complete dataset will be provided via the UCI Machine Learning library. 

\subsection{Dataset being used for training}

Note that we refer to the plays in the dataset as ``samples'', and the frequencies of the words appearing in those plays as ``features''. The goal here is to use these frequencies to determine the year each play was first performed in public. A standard best guess for date of best performance is also included in the dataset and is used for training our algorithm and measuring our accuracy. It is worth noting that the year a play was first performed is usually earlier than or the same as its year of publication, but need not be{: a play may be published before being performed}.

\begin{figure}
\centering  
\includegraphics[width=0.75\textwidth]{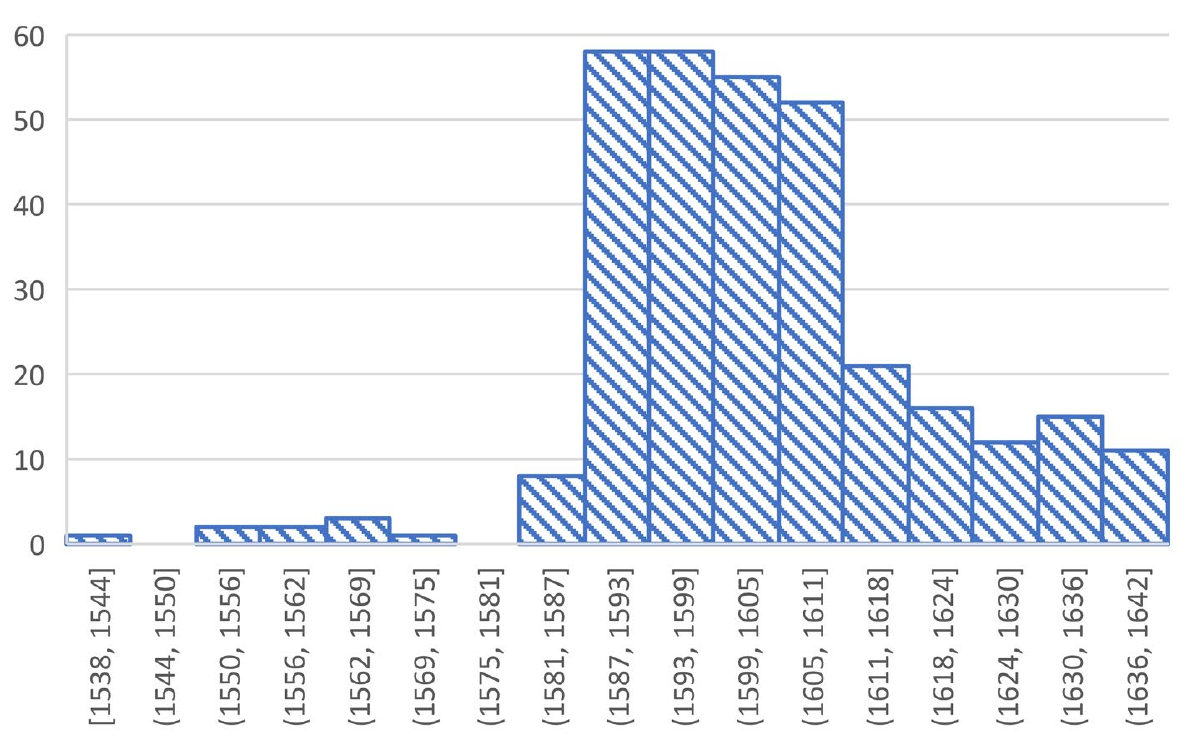}
\caption{Histogram showing the number of plays produced in each date range within the years 1538--1642. The majority of plays were first performed in the period of 1585--1610, so this is the range on which samples are extracted for training in our quest to find mathematical  models.}\label{fig:date-freq-histogram}
\end{figure}

We next examined the distribution of the plays in date ranges, to check for thinly populated ranges. We used the common formula of the square root of N to establish bin ranges, giving us seventeen bins after rounding up to the nearest integer. Four bins covering the years 1587--1611 each contained more than fifty plays, whereas none of the eight bins of earlier plays contained more than ten plays, and the best-populated of the five later bins contained just 21 plays (Figure~\ref{fig:date-freq-histogram}). We decided to concentrate on the period 1585--1610 and created a new dataset containing only the 181 plays from this date range. The new set includes 135 single-author plays by 36 individual authors, 17 multi-author plays and 29 plays of uncertain authorship. The set of word types appearing in these plays has size 51256, or 51183 if the word types that can serve as different parts of speech are each counted once rather than counted once for each of those functions. As the dataset contains a large number of features, we have to apply some feature selection methods to reduce the dimensionality of the data. We chose to use the full range of words available, avoiding the exclusion of `stop words' that is common in text mining to economise on computer resources and focus on rarer words. We used dictionary-type headwords, including inflected forms, rather than lemmas, in order to retain the extra stylistic information they carry.

\subsection{A note on feature selection} 
Our task is to find a vector -- a summed weighting of the selected features we count -- which gives the closest approximation to the date variable with the smallest possible set of features or word-variables. Since the binary Min $k$-Feature Selection problem is NP-complete and also $\W$[2]-complete~\cite{COTTA2003686}, it is unlikely that either a polynomial-time algorithm or a fixed-parameter tractable algorithm can be found for this problem. This  means that the selection of optimal sets of features for multivariate regression analysis needs to be done with some other external heuristic technique that selects them, iteratively trying combinations that lead to regression models with a progressively closer fit. Two approaches used in this study are discussed next{:} Lasso regression and the memetic algorithm for continued fraction regression.

\subsubsection{Lasso Regression}\label{sub:lasso-feature-selection}

The subset of words chosen to be included in the model was determined using the lasso regression analysis. The lasso is a well-known regularization technique for linear regression that identifies a sparse set of features. Given a linear model $y = X\beta$, where $y$ is the dependent variable, $X$ is a matrix with each column being an independent variable, and $\beta$ is the vector of parameters, along with a regularization parameter $\lambda$, lasso regression minimizes the following objective function:

\begin{equation}
    \Vert y - X\beta \Vert_2^2 + \lambda \Vert \beta \Vert_1.
\end{equation}

To ensure stability, a lasso regression model was fitted on a random subset of the data containing 80\% of the samples, repeated over 100 independent trials. The words that appear in the lasso models are shown in Table~\ref{tab:lasso_words}. A regularization parameter $\lambda $ = 1 was used.

Each word that appears in at least 90\% of lasso trials is in the top 50 words whose frequency of occurrence has the highest Pearson correlation with performance year. Thus, these 50 words  are a useful subset of features to deploy in further analysis.

\begin{table}
\begin{center}
\caption{A table containing all 14 words that appeared in at least one of 100 lasso regression trials using the dataset containing the chosen 181 plays and all 16383 words. The number in parentheses is the percentage (pct.) of trials each word appeared in at least once.}
\label{tab:lasso_words}

 \footnotesize{
\begin{tabular}{lrlrlr} 
\hline
 word & pct. & word & pct. & word & pct. \\ 
 \hline
 `and' & (100\%) & `a' & (100\%) & `you' & (100\%) \\ 
 `thou' & (100\%) & `it' &(100\%) & `is' &(99\%) \\ 
 `your' & (99\%) & `thy' &(96\%) & `sir' &(56\%) \\ 
 `my' & (15\%) & `that[conjunction]' & (14\%) & `to[infinitive]'& (8\%) \\ 
 `the' & (7\%) & `of' & (1\%) & {} & {} \\ 
 \hline
\end{tabular}
}
\end{center}
\end{table}

\subsubsection{Continued Fraction Regression}

In 2019, a regression approach based on `Continued Fraction' (\texttt{CFR}) was proposed; it views multivariate regression as a non-linear optimization problem and the authors used
a memetic algorithm to find approximations to the unknown target functions from experimental data~\cite{DBLP:conf/cec/SunM19}. Memetic algorithms are a population-based approach to solve computational problems that are posed as optimization tasks and have been heavily used for other data analytics in combinatorial optimization problems~\cite{DBLP:journals/memetic/GabardoBM20,DBLP:journals/ci/ZaherBNM19,DBLP:journals/join/HaqueM19}
and that are also showing impressive results for non-linear regression problems~\cite{moscato2019analytic, DBLP:conf/cec/SunM19,DBLP:conf/cec/MoscatoSH20,2020arXiv201203774M}
and other machine learning problems~\cite{DBLP:books/sp/19/MoscatoM19}. 

Continued fractions are a type of mathematical expression consisting of the sum of an integer and a quotient, whose denominator is again the sum of an integer and a quotient. These expressions may be finite or infinite~\cite{sun}. Euler's continued fraction formula allows us to write the sum of products as a continued fraction, as follows.
$$ x = a_0 + a_0a_1 + a_0a_1a_2 + \ldots + a_0a_1a_2\dots a_n = \cfrac{a_0}{1 - \cfrac{a_1}{1 + a_1 - \cfrac{a_2}{1 + a_2 - \cfrac{\ddots}{\ddots \frac{a_{n-1}}{1 + a_{n-1} - \frac{a_n}{1+a_n}}}}}}.$$
This simple yet powerful equation displays a general continued fraction approximation for the ratio of two higher-order polynomials. We can use the same idea to approximate a function $f(\boldsymbol{x})$ by replacing each $a_i$ and $b_i$ with other functions of $\boldsymbol{x}$. 
\cite{DBLP:conf/cec/SunM19} proposed that we can approximate the ``target function'' of a multivariate regression problem, given a set of examples, and that it can be expressed as a multivariate function $f : \mathbb{R}^n \to \mathbb{R}$ of the form:

\begin{equation}
f(\mathbf{x}) = 
g_0(\mathbf{x})\, + \, \cfrac{h_0(\mathbf{x})}{g_1(\mathbf{x}) + \cfrac{h_1(\mathbf{x})}{g_2(\mathbf{x}) + \cfrac{h_2(\mathbf{x})}{g_3(\mathbf{x}) + \ddots}}}
\label{general-equation-for-CF}
\end{equation}
Then we have $g_i(\mathbf{x}) \in \mathbb{R}$ for all integer $i\geq 0$, and each
function $f_i : \mathbb{R}^n \to \mathbb{R}$ is associated with a different array $\mathbf{a_i} \in \mathbb{R}^n$ and with a different constant $\alpha_i \in \mathbb{R}$:

\begin{equation}
g_i(\mathbf{x}) = 
\mathbf{a_i}^\mathrm{T} \mathbf{x} + \alpha_i,
\label{eqn:for-g}
\end{equation}
For each function $h_i : \mathbb{R}^n \to \mathbb{R}$ we also have a different 
array $\mathbf{b_i} \in \mathbb{R}^n$ as well as a different constant $\beta_i \in \mathbb{R}$:
\begin{equation}
h_i(\mathbf{x}) = 
\mathbf{b_i}^\mathrm{T} \mathbf{x} + \beta_i.
\label{eqn:for-h}
\end{equation}

The ``depth'' of a continued fraction refers to the number of ``subfractions'' in the overall fraction. For example, the depth 0 form of the fraction in Equation~\eqref{general-equation-for-CF} would be $x = g_0(\mathbf{x})$, the depth 1 form would be $x = g_0(\mathbf{x}) + \frac{h_0(\mathbf{x})}{g_1(\mathbf{x})}$, and so on.

It is often useful to represent continued fractions in a way that explicitly states each numerator and denominator, particularly when a continued fraction is difficult to visualize in the standard representation. To do this, we simply state the expression for each $g_i(\mathbf{x})$ and $h_i(\mathbf{x})$ term. To illustrate this, we will use the concrete example of the Mills ratio. This value is used in probability and its definition is shown in Equation~\eqref{eq:mills-ratio}, where $D(\mathbf{x})$ and $P(\mathbf{x})$ are the distribution and probability density functions, respectively~\cite{mills_wolfram}.
\begin{equation}
    m(\mathbf{x}) = \frac{1 - D(\mathbf{x})}{P(\mathbf{x})}
    \label{eq:mills-ratio}
\end{equation}

This quantity can be approximated by the following continued fraction, which appears in the two equivalent forms in Equation~\eqref{eq:mills-approx-cont-frac-1} and Equation~\eqref{eq:mills-approx-cont-frac-2}~\cite{gasull2014mills}. We will use both representations throughout this paper.
\begin{equation}
    f(\mathbf{x}) = 0 + \cfrac{1}{\mathbf{x} + \cfrac{1}{\mathbf{x} + \cfrac{2}{\mathbf{x} + \cfrac{3}{\mathbf{x} + \ddots}}}}
    \label{eq:mills-approx-cont-frac-1}
\end{equation}

\begin{align}
  \begin{split}
  g_0(\mathbf{x}) &= 0 \qquad  h_0(\mathbf{x}) = 1 \qquad g_1(\mathbf{x}) = \mathbf{x}\\
  h_1(\mathbf{x}) &= 1 \qquad g_2(\mathbf{x}) = \mathbf{x} \qquad h_2(\mathbf{x}) = 2\\
  g_3(\mathbf{x}) &= \mathbf{x} \qquad h_3(\mathbf{x}) = 3 \qquad g_4(\mathbf{x}) = \mathbf{x}  + \ddots
  \end{split}
  \label{eq:mills-approx-cont-frac-2}
\end{align}

In situations like the one we are addressing in this study, finding a multivariate regression of a single target variable, we need to approximate the unknown target function
given a dataset $S=\{(\mathbf{x^{(i)}}, y^{(i)})\}$, i.e. a set of pairs of an unknown multivariate target function $f : \mathbb{R}^n \to \mathbb{R}$ on which the image values are known (ideally, with no uncertainties). In general, 
better generalization outcomes are expected if we identify the subset of the variables of $\mathbf{x}$, which are more relevant for prediction. Minimization of the MSE on the values of the training set $S$ are used to identify the  sets of coefficients $\{\mathbf{a_i}\}$, $\{\mathbf{b_i}\}$, $\{\alpha_i\}$, and $\{\beta_i\}$. 
One of the advantages of our method is that, since it selects subsets of variables as well as adapting the coefficients in the formula, it may lead to insights about the classes of variables that are more relevant for prediction. We are going to utilize that advantage of CFR in this contribution.

\subsection{Memetic Algorithm for Iterative Continued Fraction Regression}

Memetic Algorithms (MAs) are a type of population-based approach used for solving complex problems which are generally posed as an optimization task with one or multiple objectives and constraints. 
In these methods we start by initializing the search using a ``population'' of potential solutions (generally feasible solutions of the problem at hand), which are evaluated based on some heuristic (such as mean squared error, or MSE). The fittest ones, according to this heuristic, are modified and combined to generate a new population of solutions, for which this process is repeated. MAs then follow similar process to other evolutionary type of algorithms and heuristics but they are characterized by the inclusion of an additional step of individual optimization. Each solution is then independently improved using the given heuristic before the ``recombination'' operation processes them. This increases, on average, the accuracy of solutions as well as the diversity of the new generation~\cite{neri,eorms0515}.

The CFR algorithm has a number of default parameters, which we will describe here. Unless stated otherwise, these were the parameters used for each experiment. No normalization is done on the data. The objective function, used to measure the accuracy of each potential solution, is the MSE by default. The penalty in the fitness function (the ``delta''), is 0.10 with this dataset. A larger value of this parameter prevents overfitting to the data in order to accommodate outliers. The program runs for 200 generations, where each generation is a new population of the potential solutions. The mutation rate is 0.10, which affects how much the potential solutions are altered at each stage. The root of the population tree, which determines which potential solutions will be generated, gets reset if the MSE does not improve after five generations. The local search algorithm, to improve current potential solutions, is performed at each generation. At the local search step, the Nelder-Mead algorithm is run four times, with each run producing at most 250 generations, and the algorithm resets after ten consecutive generations without improvement. Local search optimization is run serially. All data samples are used in the local search.

The depth of the continued fraction solution generated begins at 0. Once we have the depth 0 solution (using a random function as its initial solution), we use that as the initial solution to find a new solution of depth 1. This process is repeated until we reach a solution with MSE worse than that at the previous depth. At this point, we take the solution of the previous depth to be our final solution. This approach of iteratively increase the depth of the CFR algorithm as long as the fitness improves is referred to as iterative continued fraction regression ({abbreviated to} \texttt{iter-CFR}).

\subsubsection{An univariate example of the performance of the Memetic Algorithm for regression using continued fractions}

\begin{figure}
\centering
\includegraphics[width=0.7\textwidth]{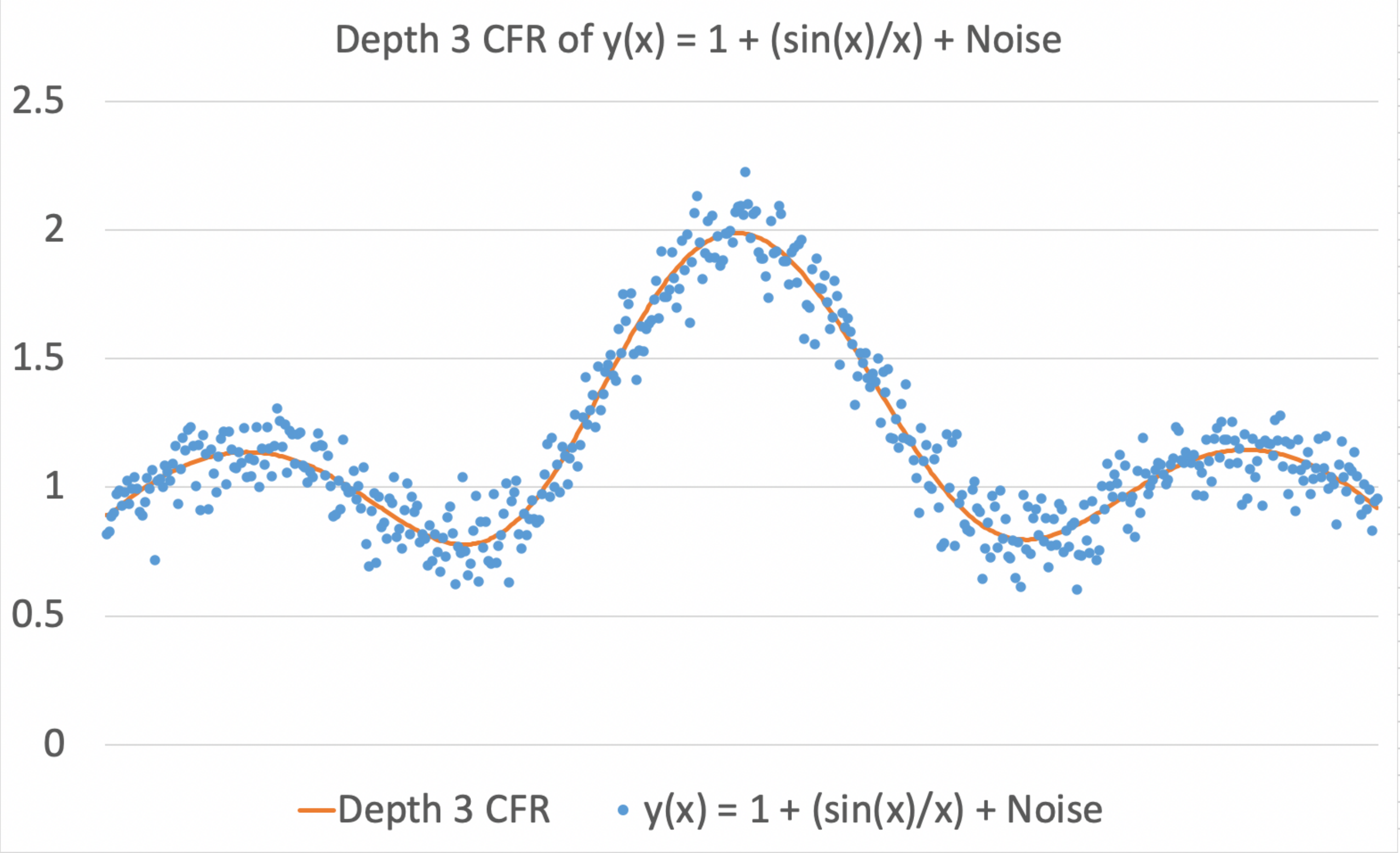}
\caption{Model produced by CFR algorithm at depth 3 on a benchmark dataset produced by adding noise to $y(x)=1+\frac{\sin{x}}{x}$ on 500 points equally separated in the interval $x\in [-10,10]$. The noise was normally distributed with mean 0 and standard deviation 0.1. The memetic algorithm found a truncated continued fraction approximation of $y(x)$ (i.e. $f(x)$ as given by Equation~\ref{general-equation-for-CF}) having a Mean Squared Error of 0.00968963.}\label{fig:results-on-sinc}
\end{figure}

As an example of the power of the memetic algorithm to do a regression of non-linear functions, we show results on approximating an unknown highly non-linear target function, namely $1+Sin(x)/x$ on the interval $[-10,10]$ and with an added normally distributed random noise with mean 0 and standard deviation of 0.01. Figure~\ref{fig:results-on-sinc} shows the approximation found with the memetic algorithm and a continued fraction of depth equal to three. We have instructed the algorithm to make use of the original variable $x$ and the metafeature $x^2$. 

Using the notation for $f(x)$ given by Equation~\ref{general-equation-for-CF}, we can then write:
\begin{align}
  \begin{split}
g_0(x) &=1.29492-0.0162327 \: x^2, \quad h_0(x) =33.8386-4.84268 \: x^2, \\
g_1(x) &=16.4545+0.580148 \: x-3.54912 \: x^2, \quad h_1 =-98.6612-3.87476 \: x-17.5014 \: x^2, \\
g_2(x) &=-6.07812-0.0996804 \: x^2, \quad h_2(x) =51.4633-0.00939706 \: x+2.38741 \: x^2, \\
g_3(x) &=16.9629-0.134414 \: x^2. 
\end{split}
\end{align}

\section{Experiments with 11 regression techniques well-known in machine learning}

To test the performance of many machine learning algorithms, we employed a dataset consisting of 181 plays and, as variables, the percentages of occurrences of the 50 words having the highest Pearson correlation with performance year of the play (ranging from 1585 to 1610). To ensure reproducibility, we employed the implementations of 11 machine learning regression methods -- comprising a set of 9 regressors from the popular \textit{Scikit-learn} machine learning library~\cite{scikit-learn}, one from \textit{XGBoost}~\cite{Chen:2016:XGBoost} and the iterative Continued Fraction presented in this paper -- to predict the year using 100 randomized runs with 80-20 training/test splits. The names of the regression methods studied are:
    {AdaBoost} (\texttt{ada-b}),
    {Gradient Boosting} (\texttt{grad-b}),
    {Kernel Ridge} (\texttt{krnl-r}),
    {Lasso Lars} (\texttt{lasso-l}),
    {Linear Regression} (\texttt{l-regr}),
    {Linear SVR} (\texttt{l-svr}),
    {MLP Regressor} (\texttt{mlp}),
    {Random Forest} (\texttt{rf}),
    {Stochastic Gradient Descent} (\texttt{sgd-r}) ,
    {XGBoost} (\texttt{xg-b}) and
    {Iterative Continued Fraction Regression} (\texttt{iter-CFR}).

In our initial testing we found that \texttt{krnl-r, l-svr, mlp} and \texttt{sgd-r} performed poorly in terms of the MSE score. We used \texttt{`squared\_epsilon\_insensitive'} loss function for SGD Regressor (with \texttt{learning\_rate=`adaptive'}) and  Linear SVR. This loss function applies the squared penalty by ignoring any residuals $(y-p) > \epsilon$ and linear in the other case. It is computed as $Loss = max\{ 0, |y - p|-\epsilon \}^2$, where $\epsilon=0.1$, $y$ and $p$ are the actual/target and predicted value. For the Kernel Ridge, we used the \texttt{`polynomial'} kernel with degree 3. As the solver in \texttt{mlp} and \texttt{sgd-r} were not converged with default parameter value for maximum iteration, \texttt{`max\_iter'}, we set the value as 25000 and 100000, respectively. We kept the default parameters of other machine learning regression algorithms unchanged.

Table~\ref{tab:100-runs-summary-stat} shows the descriptive statistics of the MSE scores obtained for 100 runs by 11 regressors. Here, we can see that \texttt{grad-b} obtained the best average MSE score of 0.08 for the training data. The best average testing MSE value of 15.65 is obtained by \texttt{ada-b}. However, \texttt{grad-b, l-regr, rf} and \texttt{xg-b} also obtained nearly the same value of average MSE score in testing (ranging from 16.03 to 16.58). The \texttt{iter-CFR} is the next closest method to that group of regressors in terms of the average test MSE of 21.25. Some other regressors performed significantly worse.

\begin{table}
\centering
        \caption{Descriptive Statistics for the 100 runs of 11 regressors on the 50 most correlated words of 181 plays  with 80-20 training/test splits.}
    \label{tab:100-runs-summary-stat}
     \footnotesize{

    \begin{tabular}{lrrrrrr}
\hline
\multirow{2}{*}{Regression Method} &  \multicolumn{3}{c}{Training MSE Score} & \multicolumn{3}{c}{Testing MSE Score} \\
\cline{2-7}
 &     Avg.         &     Med.         &      std.        &       Avg.       &     Med.         &       std.       \\
\hline

ada-b     &         2.88 &         2.85 &         0.25 &        \textbf{15.49} &        \textbf{15.65} &         4.07 \\
grad-b    &         \textbf{0.09} &         \textbf{0.08} &         \textbf{0.02} &        16.28 &        16.22 &         4.24 \\
iter-CFR  &        12.14 &        11.93 &         1.91 &        21.25 &        20.41 &         8.56 \\
krnl-r    &      1722.95 &      1726.55 &        28.72 &      1974.71 &      1908.89 &       559.22 \\
l-regr    &         6.13 &         6.17 &         0.56 &        16.56 &        16.54 &         3.97 \\
l-svr     &      1830.26 &      1835.50 &        31.53 &      2545.38 &      2351.11 &       781.39 \\
lasso-l   &        47.99 &        48.01 &         1.54 &        47.67 &        47.52 &         6.04 \\
mlp       &         5.69 &         5.46 &         1.91 &        73.91 &        61.55 &        42.60 \\
rf        &         2.23 &         2.24 &         0.18 &        15.93 &        15.77 &         4.10 \\
sgd-r     &       266.58 &       197.66 &       280.18 &       520.90 &       369.56 &       529.74 \\
xg-b      &         0.54 &         0.55 &         0.08 &        16.11 &        15.89 &         \textbf{3.95} \\

\hline
\end{tabular}
}
\end{table}

\begin{figure}
\centering
\includegraphics[width =0.9\textwidth]{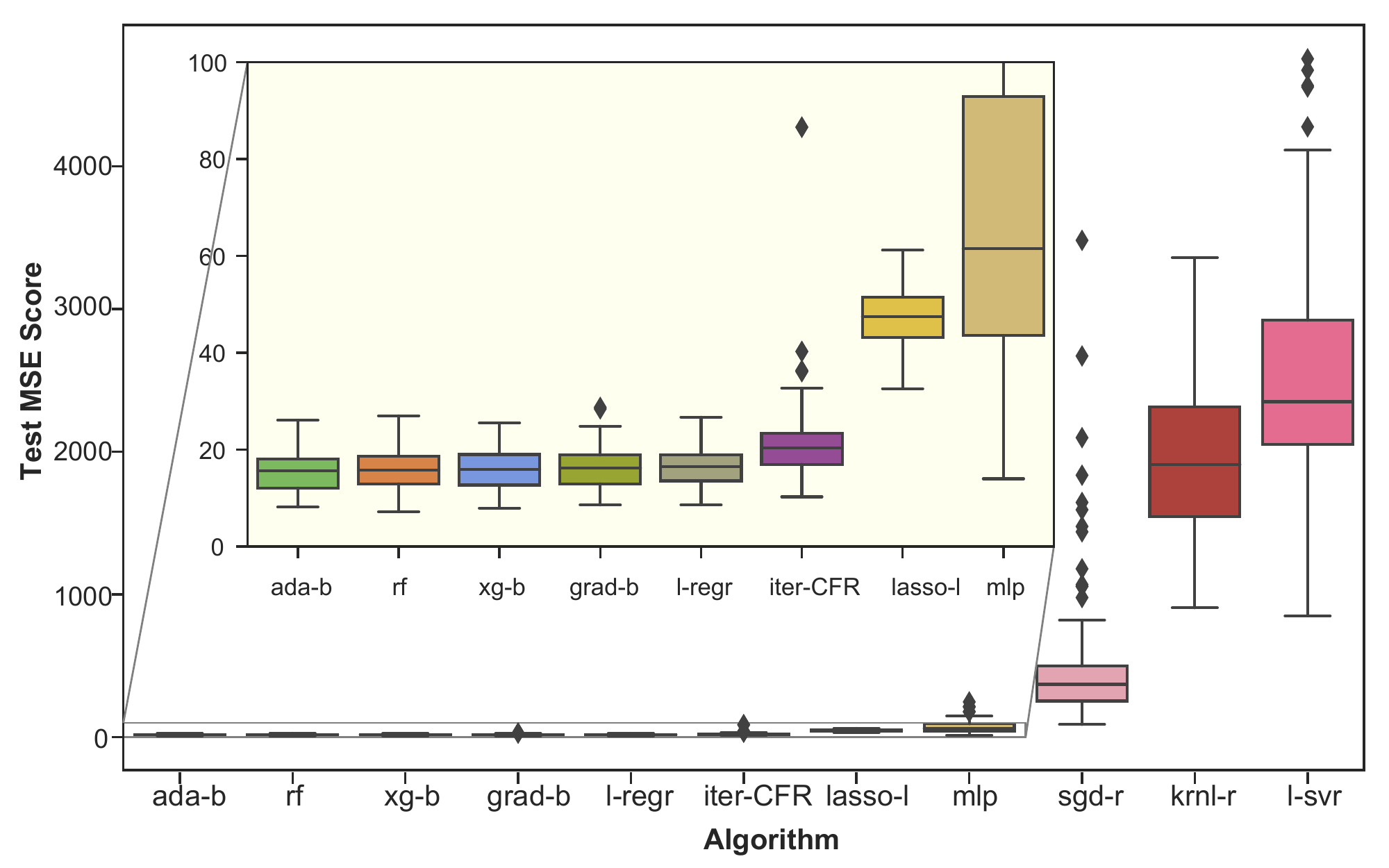}
\caption{Bar and whisker plot showing the MSE scores of regressors obtained for 100 runs in the testing sets. As the MSE scores of the regressors vary in a wide range, we show the subset of regressors having the upper bound of MSE score of 100 in the inset.}
\label{fig:box-test}
\end{figure}

In addition to the summary table, we show in Figure~\ref{fig:box-test} the box plot for the testing MSE scores of the regressors obtained for 100 runs. From this plot, it can be seen that the range of MSE scores is wide. To better understand the MSE scores obtained by a good subset of regressors, we show the zoomed plot as an inset for Test MSE scores up to 100. From the  inset we can see that \texttt{ada-b, rf, xg-b, grad-b} and \texttt{l-regr} exhibited similar results to \texttt{iter-CFR} as the closest performing regressor to the group.

\subsection{Statistical Comparison of the Rankings of Regression methods}

We conducted the Friedman test for repeated measure~\cite{friedman1937use} to validate the significance in results obtained by different regression methods for 100 independent runs. We used the ranking of the methods based on their MSE scores obtained for the test set to help us determine if the experiment's techniques are consistent in their generalization performance. The statistical test found  \num{2.74845e-176} which \textit{rejected} the \textit{null} hypothesis of \textit{all the models perform the same} and we proceeded with the post-hoc test.

\begin{figure}
\centering
    \includegraphics[width=0.65\textwidth]{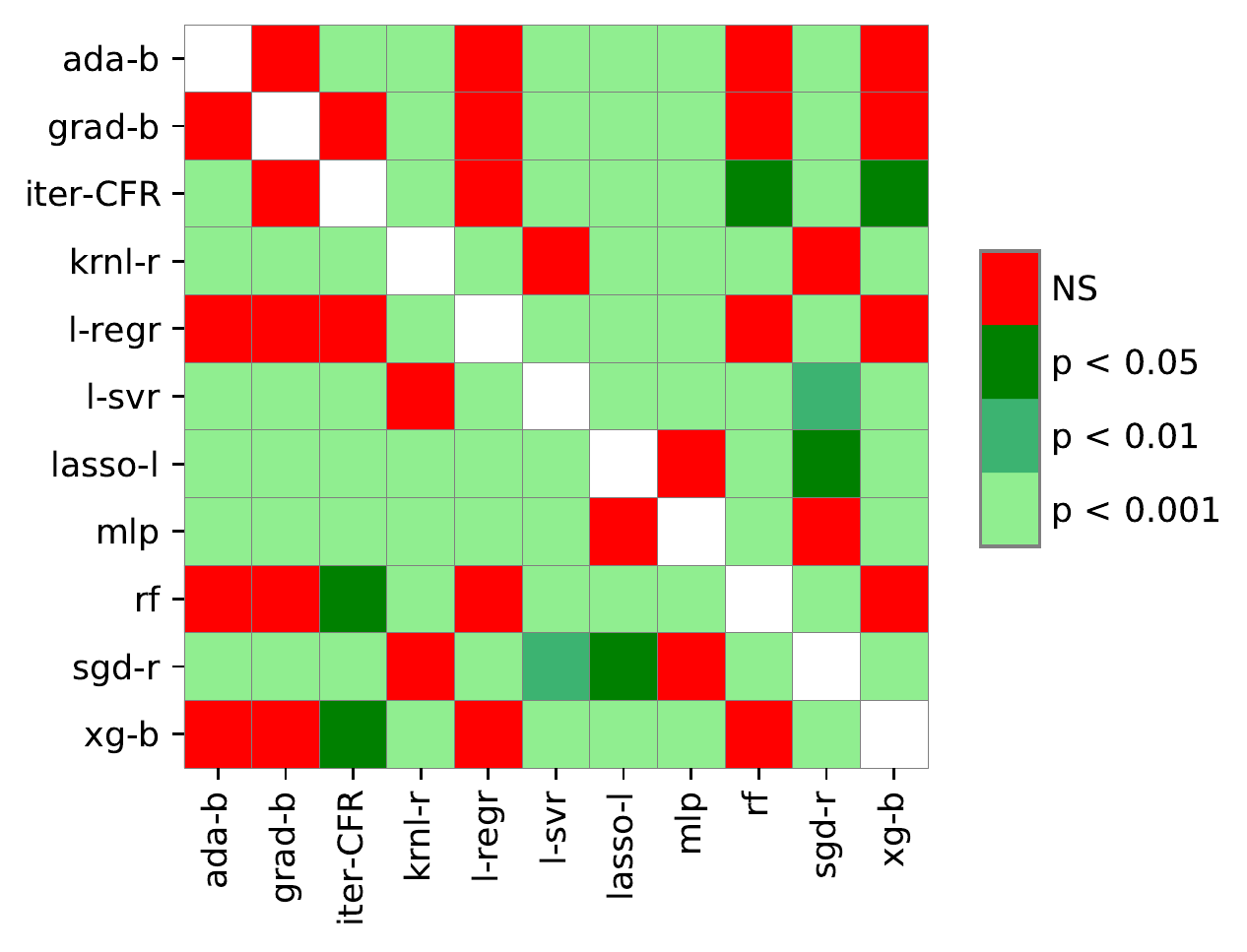}
  \caption{Heatmap showing the Statistical significance levels of $p$-values obtained by the Friedman Post-hoc Test.} 
\label{fig:heatmap-test}
\end{figure}

\begin{figure}
\centering
    \includegraphics[width=\textwidth]{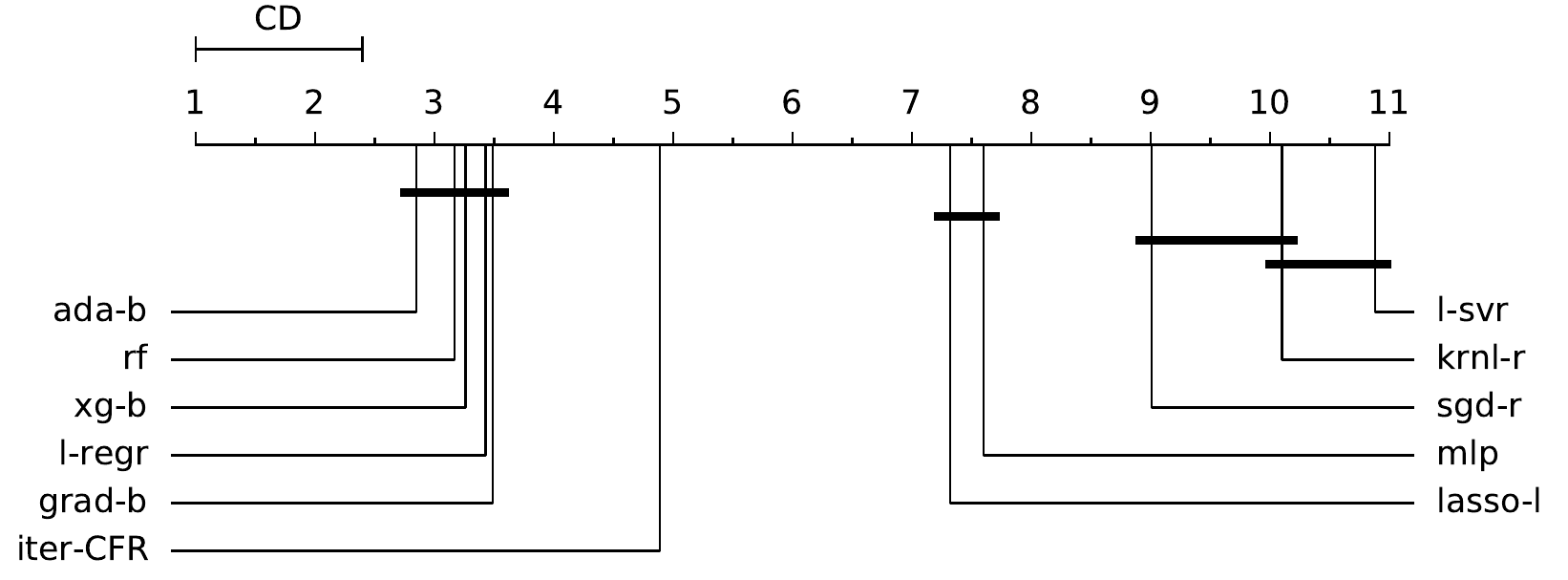}
  \caption{Critical Difference (CD) plot showing the statistical significance of rankings achieved by the regression methods.} 
\label{fig:cd-plot-test}
\end{figure}

The Friedman's post-hoc test on the ranking of 11 regressors computed for the test \textit{MSE} scores was obtained for 100 runs on 80-20 split. In Figure~\ref{fig:heatmap-test} the $p$-values obtained for the test are plotted as a heatmap. It is noticeable that there exist \textit{no significant differences} ({symbolized as} NS in Figure~\ref{fig:heatmap-test}) in performances of \texttt{iter-CFR} with \texttt{l-regr} and \texttt{grad-b}.

Additionally, we generated the Critical Difference (CD) diagram proposed by~\cite{demvsar2006statistical} to visualize the differences among the regressors regarding their median ranking. The CD plot uses the Nyemeni post-hoc test and hence the results may differ from the results obtained from the results obtained from Friedman's post-hoc test, and it places the regressors on the $x$-axis of their median ranking. It then computes the \textit{critical difference} of rankings between them. It connects those methods which are closer than the critical difference with a horizontal line denoting that them as statistically \textit{non-significant}.

In Figure~\ref{fig:cd-plot-test} we plot the CD graph using the implementation from the Orange data mining toolbox~\cite{Python:Orange} in Python.  The Critical Difference is found to be $1.397$. We can see that there are no significant differences in the median rankings of \texttt{rf, xg-b, l-regr} and \texttt{grad-b} with the top-ranked \texttt{ada-b}. The ranking of \texttt{iter-CFR} is statistically not similar to any other regressors; however, it is next to the group of regressors that ranked first.

\section{Discussion}
\begin{table}
\centering
\caption{Number of times each regression method came first and the value of maximum and minimum ranking achieved for 100 runs on the dataset.}
\label{tab:method-1st-max}
 \footnotesize{
\begin{tabular}{lrclrc}
\hline
{Regressor} &  \#\nth{1} &  Rank (min, max) & {Regressor} &  \#\nth{1} &  Rank (min, max)\\
\hline
l-regr	&	32	&	(1, 6)	&	mlp	&	0	&	(2, 9)	\\
ada-b	&	19	&	(1, 6)	&	lasso-l	&	0	&	(6, 8)	\\
grad-b	&	14	&	(1, 6)	&sgd-r	&	0	&	(8, 11)		\\
xg-b	&	13	&	(1, 6)	&	krnl-r	&	0	&	(9, 11)	\\
rf	&	11	&	(1, 6)	&	l-svr	&	0	&	(10, 11)	\\
iter-CFR	&	11	&	(1, 7)	\\						
\hline
\end{tabular}
}
\end{table}

Table~\ref{tab:method-1st-max} shows the number of times a regressor was ranked first and its minimum and maximum ranking for 100 runs on the dataset. We can see that the same set of regressors, the ones that are ranked  first in the CD plot of Figure~\ref{fig:cd-plot-test}, have a maximum ranking of 6. In terms of the number of times a regressor was ranked \nth{1}, \texttt{l-regr} has the highest value (32 times). The proposed \texttt{iter-CFR} was ranked first 11 times in 100 runs, which is same as the value for the regressor \texttt{rf}. The set of regressors consisting of \texttt{krnl-r, l-svr, lasso-l, mlp} and \texttt{sgd-r} were never ranked first in the 100 runs. Moreover,  \texttt{l-svr} exhibited the worst ranking with the minimum ranking of 10 out of 11 regressors.

\begin{table}
    \centering
    \caption{20 most frequent words showing the number of times ($x$) each has appeared in 100 models of iterative Continued Fraction Regression (\texttt{iter-CFR}) using the 50 words whose frequencies are most correlated with date.}
    \label{tab:20-freq-words}
    \footnotesize{
    \begin{tabular}{lrlrlrlr}
    \hline
    word & $x$ & word & $x$ & word & $x$ & word & $x$ \\
    \hline
    `ah' & 66 & `that[conjunction]' & 54 & `own' & 49 & `business' & 35 \\
    `goodness' & 60 & `your' & 53 & `women' & 48 & `does' & 34 \\
    `known' & 59 & `beseems' & 52 & `aside' & 38 & `wherein' & 32 \\
    `therefore' & 56 & `for[conjunction]' & 52 & `content' & 38 & `thy' & 25 \\
    `like[preposition]' & 55 & `has' & 50 & `thou' & 36 & `threats' & 25 \\
    \hline
    \end{tabular}
    }
\end{table}

\subsection{Looking in depth at the best model of iterative continued fraction}

\begin{figure}
\centering 
    \subfloat[Training  Performance]{\includegraphics[width=0.49\textwidth]{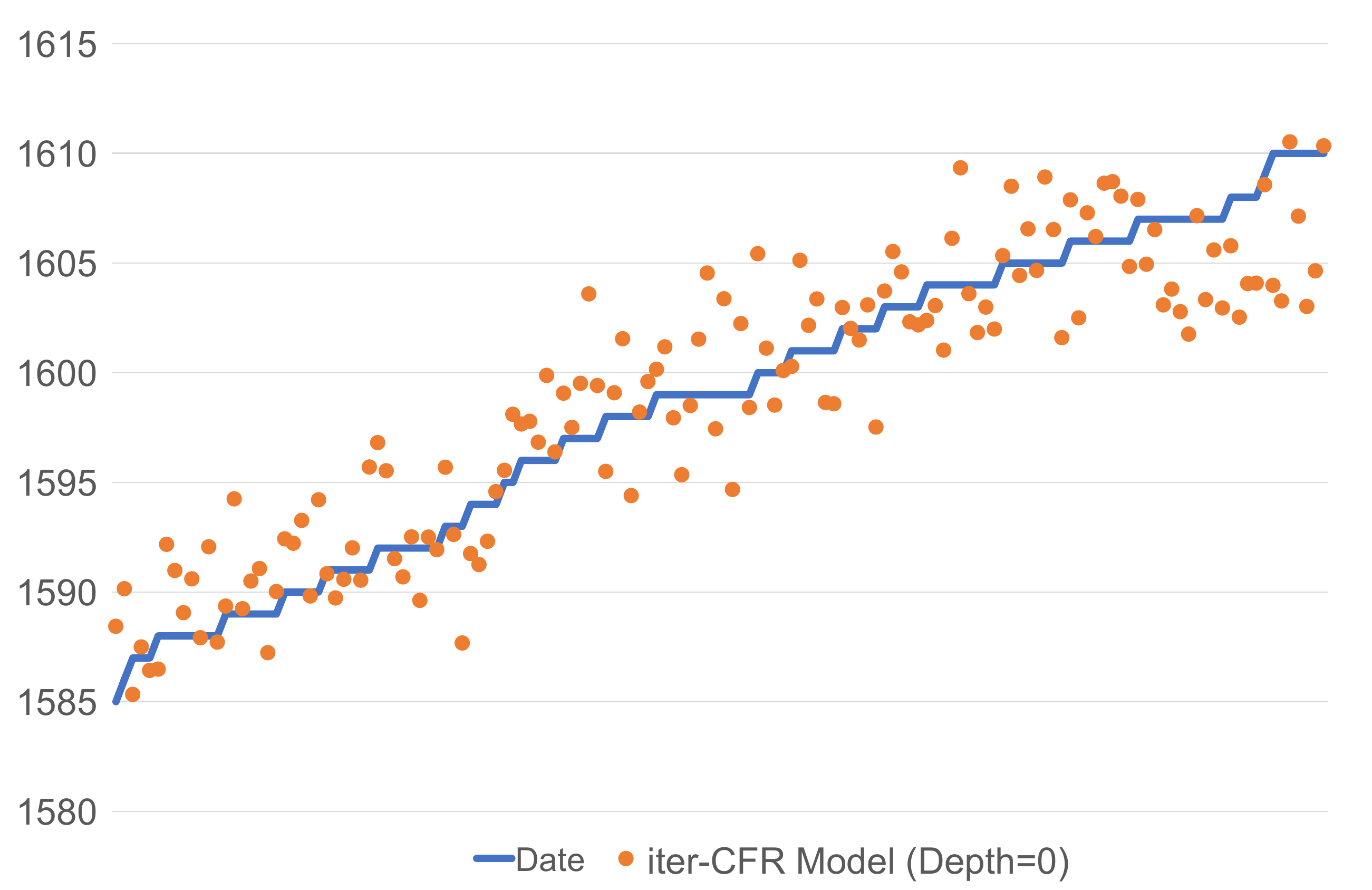}}\label{fig:iter-cfr-train}
    \hfill
    \subfloat[Testing Performance]{\includegraphics[width=0.49\textwidth]{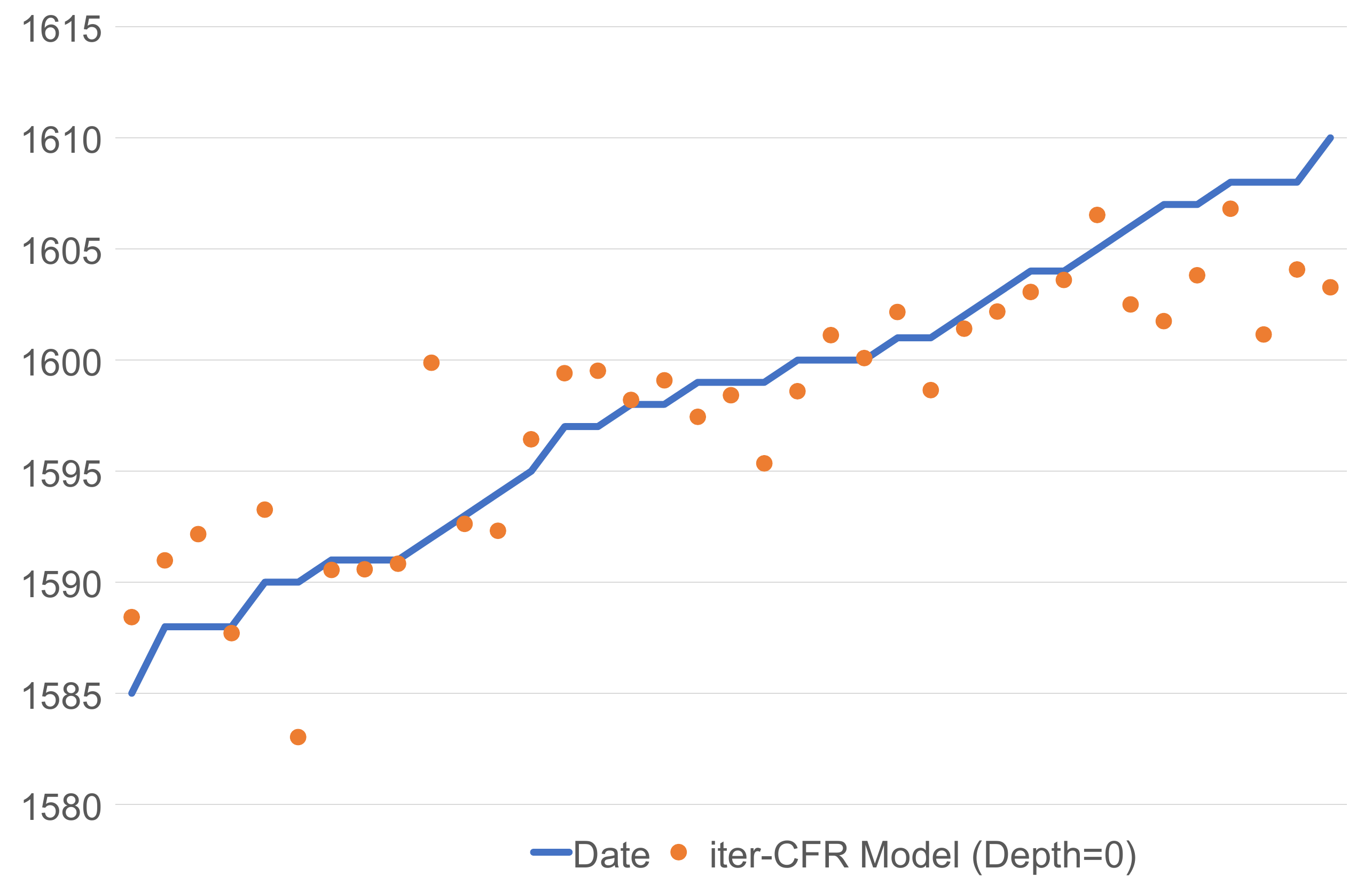}}\label{fig:iter-cfr-test}
    \\
  \caption{The result of the best model found for iterative CFR algorithm on the dataset containing only the top 50 words whose frequencies of occurrence have the highest Pearson correlation with performance year. The blue function is the target, and the orange dots are the approximation. a) Result of the CFR algorithm at depth 0 on the training portion of the data with MSE score of 7.637. b) Result of the CFR algorithm at depth 0 on the testing portion of the data with MSE score of 14.475.} 
\label{fig:iter-cfr-fit}
\end{figure}

We look at the best \texttt{iter-CFR} model found in the 100 repetitions of the experiment. The model which fitted the training data best had a training MSE of 7.63661 and produced a test MSE of 14.4752. The continued fraction model is at depth=0 and it is as follows:

\begin{equation*}
\begin{split}
f(x) &= 1608.8+13.0358\times (has) -16.5958\times (ah)-7.10359\times (for[conjunction])\\
&-7.00464\times (that[conjunction])-4.7321\times (thy)+54.2334\times (known)\\
&-400.11\times (beseems)+31.6509\times (women)-104.339\times (wherein) \\
&-104.749\times (aside)+1.39045\times (that[demonstrative])+1.27499\times (mighty)\\
&+110.119\times (goodness)-2.8458\times (a) - 68.8078\times (saith) -44.4536\times (triumph) \\
&+15.6812\times (like[preposition])-8.98319\times (words).
\end{split}
\end{equation*}
12 out 18 of these words, `ah', `goodness', `known', `like[preposition]', `that[conjunction]', `beseems', `for[conjunction]', `has', `women', `aside', `wherein' and `thy' are in the 20 most frequent words (Table ~\ref{tab:20-freq-words}). In Figure ~\ref{fig:iter-cfr-fit} we show how well the continued fraction model predicts the year for both the training and the testing portions of the data.

\subsection{Top 20 Words by Interpretable Models}
\begin{figure}
\centering
\includegraphics[width=0.7\textwidth]{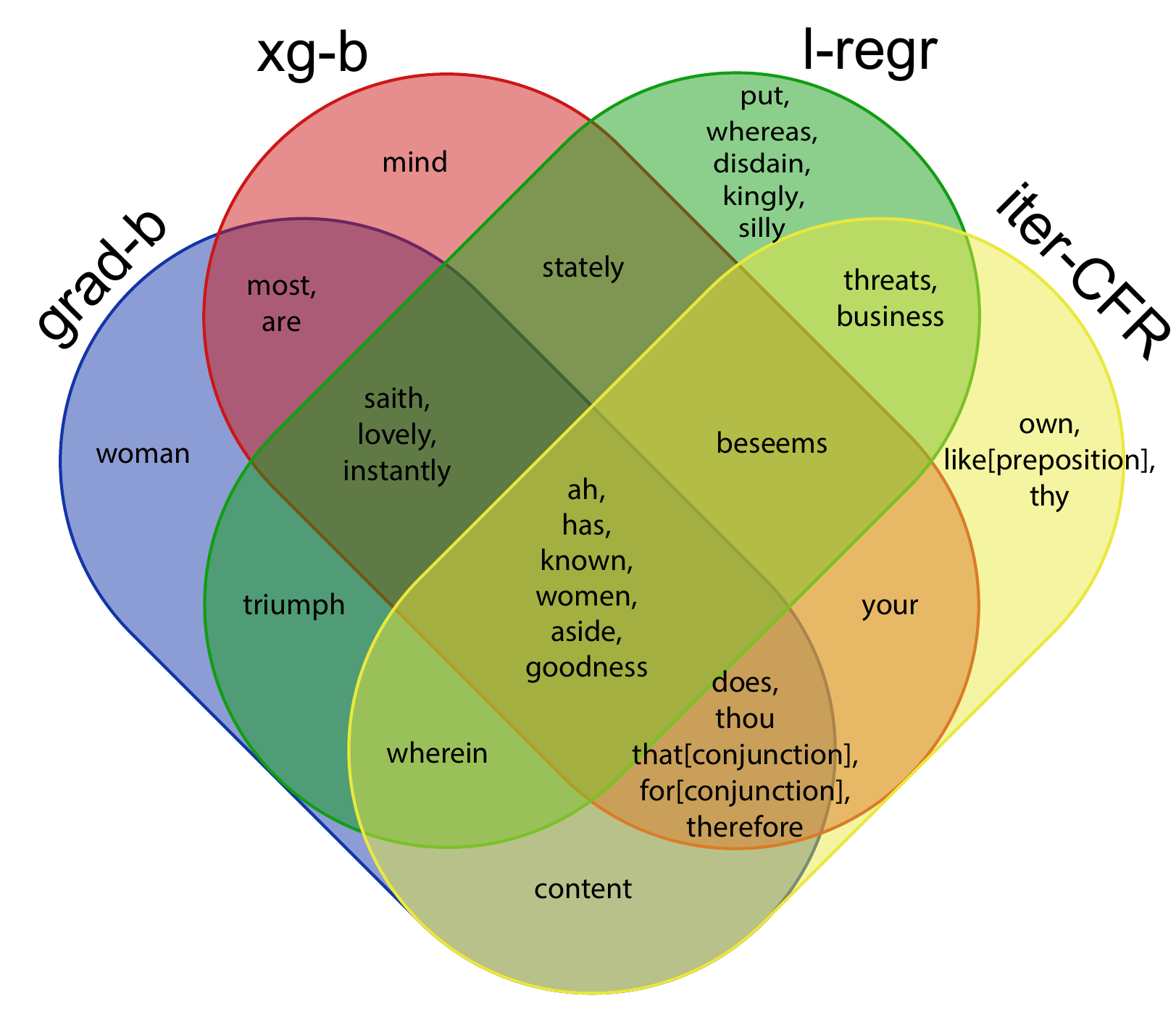}
  \caption{Venn diagram showing the agreement in top 20 words of the models by \texttt{grad-b}, \texttt{xg-b}, \texttt{l-regr} and \texttt{iter-CFR}. A subset of six words is common to all four methods: `ah', `has', `known', `women',	`aside' and	`goodness'.} 
\label{fig:venn_top_20_words}
\end{figure}

We selected three interpretable regressors (\texttt{grad-b}, \texttt{xg-b} and \texttt{l-regr}) which ranked highly in the CD plot shown in Figure~\ref{fig:cd-plot-test}. Then we collected the \textit{feature importance} score of the words given by each of their 100 models. From those scores we selected the top 20 words for each regressor and compared them with the 20 most frequently appearing words from \texttt{iter-CFR} in the Venn diagram in Figure~\ref{fig:venn_top_20_words} created with an online Venn Diagrams tool developed by Van de Peer Lab~\footnote{Venn Diagrams tool can be accessed form the Website of Bioinformatics and Evolutionary Genomics group of Ghent University, Belgium at  \url{http://bioinformatics.psb.ugent.be/webtools/Venn/}}. We can observe strong agreement in selecting words by \texttt{iter-CFR} and other regressors. Our \texttt{iter-CFR} has 13 common words with each of \texttt{grad-b} and \texttt{xg-b}, and 10 common words with \texttt{l-regr}. Among the 20 words of \texttt{iter-CFR}, only three words --- `own', `like[preposition]', `thy' --- did not appear in any of the top 20 words given by other methods. Due to these strong correspondences with the 20 most frequently appearing words found by \texttt{iter-CFR} and other regressors, we analyze these words' roles in \texttt{iter-CFR} models in the following section.

\subsubsection{The 20 Most frequently appeared words in iterative continued fraction models}

\begin{figure}
\centering 
  \includegraphics[width=0.9\textwidth]{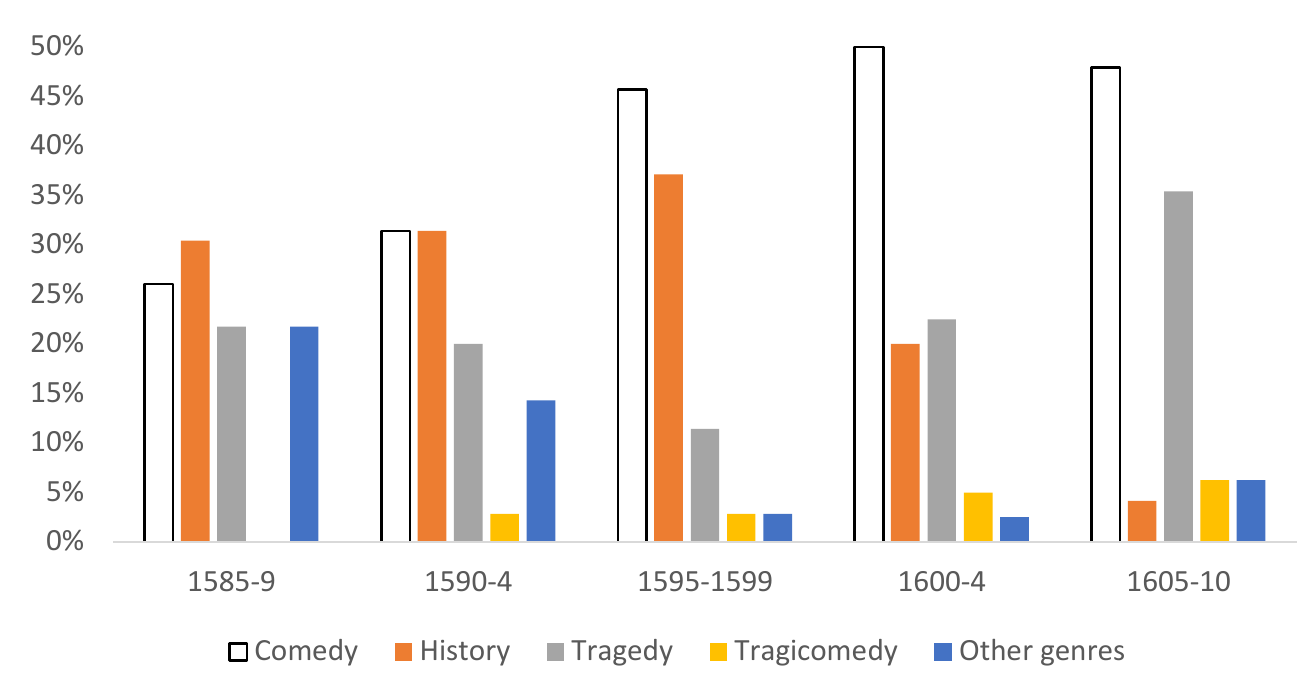}
  \caption{Percentage of plays in five genre groupings, by half- decade. Comedy is well represented throughout. History plays decline sharply after the third half-decade.} 
\label{fig:CFR-20-words-discussion-genres-by-half-decade}
\end{figure}

 Figure~\ref{fig:CFR-20-words-discussion-genres-by-half-decade} shows the percentage of plays in five genre groupings by the period of 1885--1610. ``Comedy'' is well-represented here; however, ``History'' plays decline sharply after the third half-decade. We will analyze the association of words with the genre. Table~\ref{tab:20-freq-words} lists the twenty words most often included in the functions which emerge from the CFR process. They are evidently markers of change over time in the style of the plays. The exclamation `ah' is the word-variable most commonly found in the functions, appearing 66 times. Its incidence declines over the period. It has been noticed before in discussions of word used in early modern English drama. The editors of the \textit{Encyclopedia of Shakespeare's Language} offer it as an example of the way ``certain words, meanings, structures{, etc.} are peculiar to tragedies, comedies or histories, to certain social groups --- and to specific periods''~\cite[pp.\ 1]{grant-Jonathon,MDS14177}.
They note that in Shakespeare's works `ah', which  ``signal[s] emotional distress or pity'', ``is characteristic of the histories, and occurs more than twice as densely in the speech of female characters compared with male''. This word is ``used relatively frequently by Shakespeare, compared with his contemporaries, and, despite being characteristic of the histories, is strongly colloquial in flavour, occurring densely in speech-related genres (e.g. trial proceedings)''.

\begin{figure}
\centering 
  \includegraphics[width=0.9\textwidth]{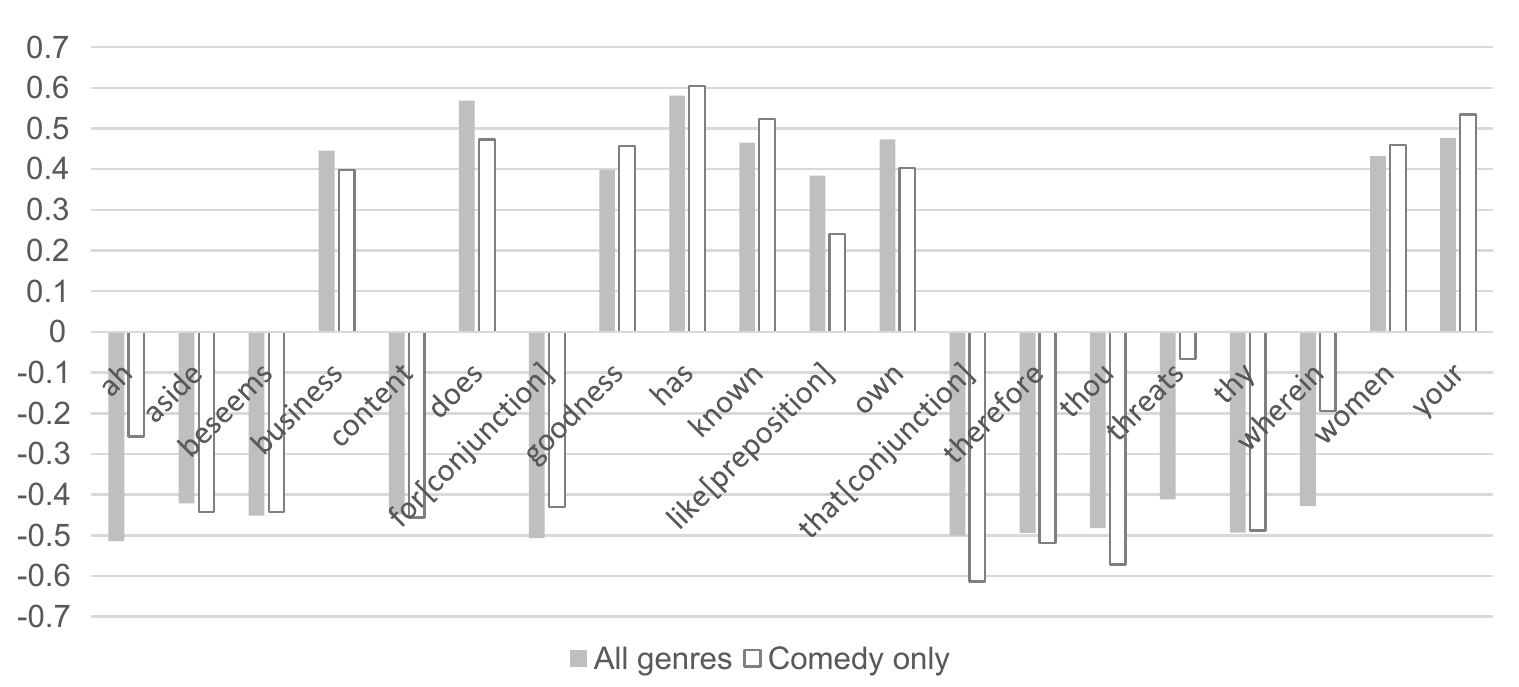}
  \caption{The Pearson product-moment correlation between date and probability for 20 word-variables found by iterative Continued Fraction Regression models in all genres and in comedies. The correlations are all significant at the $p<0.01$ level for the all-genres and comedies sets except for the correlation for the word `threats' in comedies ($r=-0.067$, $p=0.370$).
  } 
\label{fig:cfr-20-words-discussion-correlation}
\end{figure}

From our study, we can add to this that usage declines in play dialogue in general over the period 1585--1610. As we have seen, the Encyclopedia editors comment that `ah' is unusually common in Shakespeare's history plays, and we might infer that the change in usage over time can be explained by the fact that in the drama more generally, as well as in Shakespeare's canon, history plays cluster in the years before 1600, but if we account for the genre effect associated with history plays by looking exclusively at comedies (Figure~\ref{fig:cfr-20-words-discussion-correlation}), there is still a significant negative correlation between date and the probability of `ah' ($r=-0.257$, $p=0.0005$). For this purpose the genre of comedy includes plays described in Wiggins and Richardson as ``Classical Legend (Comedy)'', ``Domestic Comedy'', and ``Romantic Comedy'', as well as simply ``Comedy''.  If we include in a broader History Play category plays described in Wiggins and Richardson as ``Biblical History'', ``Classical History'', ``Legendary History'' and ``Pseudo-history'', as well as simply ``History'', we get the following percentages of History Plays compared to all plays by half-decade: 1585--89, 30.4\%; 1590--94, 31.4\%; 1595--99, 37.1\%; 1600--04, 20\%; 1605--10, 4.2\%.

A number of the words in Table~\ref{tab:20-freq-words}  are already well known as forms whose incidences were increasing or decreasing in the English language in general at this time as part of overall changes in Early Modern English. The auxiliary verbs `does' and `has' are incoming forms, replacing the outgoing forms `doth' and `hath' respectively. The older forms remained current but became progressively less common. The pronouns `thou' and `thy' are outgoing forms, and the pronoun `your' is an incoming form, part of the larger change whereby `thou' forms in general lost their function as singular second person forms, and `you' forms were increasingly used in for singular as well as plural referents. 

Some other words in Table~\ref{tab:20-freq-words} which decline in use in the plays -- `beseems', `for' as a conjunction, `that' as a conjunction, and `wherein' -- sound archaic to a modern ear and it is plausible that playwrights might use fewer of them over time in the period of our study as a reflection of contemporary usage outside the theatre. 

The remaining words have not, to the best of our knowledge, been discussed in the context of language change before. Four of them decrease in incidence over the period: `aside', `content', `therefore' and `threats'.  The use of `threats' declines significantly in the full corpus ($r=-0.0412$, $p<.0001$), but does not also decline in a separate sub-corpus composed exclusively of comedies ($r=-0.0674$, $p=0.367$), and in this case we might suspect that a genre factor might best explain its power to mark change over time and hence its presence in Table~\ref{tab:20-freq-words}. The other three show a highly significant correlation between probability and date in comedies as well as in the full set.
Five words not yet mentioned increase in incidence over the period: `business', `goodness', `known', `like' as a preposition, `own', and `women'. All of them have a highly significant correlation between probability and date in the comedies sub-corpus.

\subsection{Out of Domain Performances of the model}

To test the generalization capability of the regressors, we conducted an out-of-domain test. For this purpose, we drew 80\% data uniformly at random from the set of data with date of plays within 1885--1610 range as a train set. We trained the model on these random samples of training data and tested its generalization capability on the out-of-domain test data, containing the samples outside the range of 1585--1610. This process is repeated 100 times for getting a statistically sound understanding of their performances. The descriptive summary of the regressors sorted by Testing MSE score in ascending order is shown in Table.~\ref{tab:100-runs-summary-out-domain}. Here we can see that \texttt{mlp} has shown the best generalization performances among 11 methods. Our \texttt{iter-CFR} is ranked \nth{3} for the average MSE score obtained in Test set for 100 runs among 11 regressors.


\begin{table}
\centering
        \caption{Descriptive Statistics for the 100 runs of 11 regressors trained on the 50 most correlated words with each train set consist of randomly drawn 80\% samples from 181 plays (date in 1585--1610) and tested the generalization capability on the out-of-domain (plays with a date outside of  1585--1610 range) test data. The regressors are sorted in ascending order of their average testing MSE score.}    \label{tab:100-runs-summary-out-domain}
     \footnotesize{

    \begin{tabular}{lrrrrrr}
\hline
\multirow{2}{*}{Regression Method} &  \multicolumn{3}{c}{Training MSE Score} & \multicolumn{3}{c}{Out-of-Domain Testing MSE Score} \\
\cline{2-7}
 &     Avg.         &     Med.         &      std.        &       Avg.       &     Med.         &       std.       \\
\hline
mlp	&	5.270	&	5.118	&	1.623	&	377.685	&	373.270	&	43.403	\\
l-regr	&	6.130	&	6.167	&	0.542	&	443.024	&	441.430	&	19.799	\\
iter-CFR	&	14.479	&	14.489	&	3.163	&	451.108	&	446.530	&	41.543	\\
grad-b	&	0.089	&	0.086	&	0.018	&	456.207	&	455.175	&	12.087	\\
xg-b	&	0.548	&	0.551	&	0.077	&	476.658	&	476.912	&	14.593	\\
ada-b	&	2.869	&	2.851	&	0.247	&	478.238	&	477.300	&	11.566	\\
rf	&	2.246	&	2.237	&	0.174	&	494.535	&	494.290	&	9.875	\\
sgd-r	&	244.362	&	206.296	&	305.002	&	567.663	&	524.490	&	316.154	\\
lasso-l	&	48.006	&	47.991	&	1.606	&	723.946	&	723.272	&	7.039	\\
krnl-r	&	1722.366	&	1726.545	&	29.747	&	1957.720	&	1958.517	&	145.128	\\
l-svr	&	1829.291	&	1835.390	&	32.747	&	2325.274	&	2258.324	&	232.467	\\

\hline
\end{tabular}
}
\end{table}


To understand the importance of words, we look at the best model of \texttt{iter-CFR} on the out-of-domain test. The continued fraction model is given by:

\begin{equation*}
f(\mathbf{x}) = 
g_0(\mathbf{x})\, + \, \cfrac{h_0(\mathbf{x})}{g_1(\mathbf{x})}
\label{iter-cfr-out-domain}
\end{equation*}
where
\begin{align*}
  \begin{split}
g_0(x) &=1604.17+15.4971\times (has) - 40.1605\times (therefore)-5.3539\times (thou)\\
& +22.5657\times (own)-81.8433\times (stately)-31.7315\times (mighty), \\
h_0(x) &=-75.3675+812.856\times (has)-1730.8\times (therefore)+962.143\times (own)\\
& -614.681\times (stately)-81.0375\times (mighty),\\
g_1(x) &=5.99535+4348.8\times (has)+499.345\times (therefore)-1.50877\times (thou)\\
& -130.705\times (own)-99.6367\times (stately)+157.17\times (mighty).
\end{split}
\end{align*}

Interestingly, this \texttt{iter-CFR} model is able to obtain 423.982 MSE score on the out-of-domain test set and uses only six words (`has', `therefore', `thou', `own', `stately' and `mighty'). Among these six words, only `stately' and `mighty' did not appear in top 20 words used by \texttt{iter-CFR} tested on the data with date 1585--1610 (presented in Table.~\ref{tab:20-freq-words}). 
`Stately' and `mighty' both have a strong negative correlation with date in the full set of 285 plays ($r=-0.2403$, $p<0.0001$ and $r=-0.2759$, $p<0.001$) but the correlation with date is not significant in the subset of comedies ($r=-0.290$, $p=0.7564$ and $r=-0.949$, $p=0.3087$). It is likely that some of the change over time in the probabilities of these words is linked to the replacement of high-scoring genres with lower-scoring genres in the later plays.

\section{Conclusions}
We analyzed the frequency of words most correlated with the date of publication of $181$ English plays from the sixteenth and seventeenth centuries (ranging between 1585 and 1610). We employed a set of 11 machine learning methods on the dataset of words with their frequencies to predict the date of the first performance of the plays. In our effort to learn the significance of the words during the Shakespearean era as markers for publication date, we trained each of the machine learning regression methods with an 80\% of the data samples taken uniformly at random. We tested the methods' predictive performance on the remaining 20\% of the data and repeated this process 100 times, each with a separate set of train and test samples but with the same ratio. AdaBoost, Random Forest, XGBoost, Gradient Boosting, and Linear Regression have shown the best performance in terms of predictive capability. However, most of these models are non-interpretable, in terms of the usage information of the words. The next best performing method, supported by statistical tests, is the iterative Continued Fraction Regression (\texttt{iter-CFR}), which has the advantage of offering interpretable models. We further analyzed the 20 words from \texttt{iter-CFR}, and found that those are already well known in English plays during the Shakespearean era. As an obvious finding, the word ``ah'' is the most frequently appearing in the \texttt{iter-CFR} model, which is indeed a signature word of Shakespeare for plays from the history play genre and negative correlation with comedies. 
The relatively good performance of Linear Regression in the out-of-domain test, also indicate that for the relatively short interval analysed a linear approximation provides a good generalisation capability. 
A more in-depth analysis revealed in the context of language change, that a set of words (`aside', `content', `therefore' and `threat') showed a significant decrease in usage over the period. However,  the usage of `threat' has declined over the years but opposite trends exhibited in the comedy genre. This application of machine learning methods on the frequency of words from the plays not only uncovered some interesting insights about the relationship of word frequencies with the genre but also corroborates the understanding of certain words as the signature words of William Shakespeare.

\section*{Acknowledgment}
This work was supported by the Australian Government through the Australian Research Council's Discovery Projects funding scheme (project DP160101527, DP200102364). This work has been supported by the University of Newcastle and Caltech Summer Undergraduate Research Fellowships (SURF) program. In particular, SURF Fellows J. Sloan and K. Huang acknowledge the support of Samuel P. and Frances Krown and Arthur R. Adams, respectively, for their generous donor support to their activities.

\section*{CRediT author statement}
\textbf{Pablo Moscato:} Conceptualization, Methodology, Formal analysis, Investigation, Writing - Original Draft, Writing - Review \& Editing, Supervision, Project administration, Funding acquisition.\textbf{ Hugh Craig:} Conceptualization, Methodology, Data Curation, Writing - Original Draft, Writing - Review \& Editing, Funding acquisition. \textbf{Gabriel Egan:} Data Curation, Writing - Review \& Editing. \textbf{Mohammad Nazmul Haque:} Methodology, Software, Validation, Formal analysis, Investigation, Writing - Original Draft, Writing - Review \& Editing, Visualization. \textbf{Kevin Huang:} Methodology, Software, Validation, Formal analysis, Investigation, Writing - Review \& Editing. \textbf{Julia Solan}: Methodology, Software, Validation, Formal analysis, Investigation, Writing - Review \& Editing. \textbf{Jon Corrales de Oliveira:} Methodology, Software, Validation, Formal analysis, Investigation, Writing - Review \& Editing.

\bibliographystyle{unsrt}  
\bibliography{arXiV-IterCFR-Shakespeare}  






\end{document}